\documentclass{article}
\usepackage[utf8]{inputenc}
\usepackage{amsmath}
\usepackage{amsfonts}
\usepackage{amssymb}
\usepackage{graphicx}
\usepackage{listings}
\usepackage[hidelinks]{hyperref}
\usepackage{url}
\usepackage{booktabs}
\usepackage{multirow}
\usepackage{array}
\usepackage{enumitem}
\usepackage{courier}
\usepackage{xcolor}

\usepackage{cite}

\definecolor{codebg}{rgb}{0.95, 0.95, 0.95}

\title{Internalized Self-Correction for Large Language Models}
\author{Nishanth Upadhyaya\footnote{Freshworks, email:~\url{nishanthu@gmail.com}}~~and Raghavendra Sridharamurthy\footnote{Indian Institute of Science, email:~\url{g.s.raghavendra@gmail.com}}}
\date{}

\begin{document}

\maketitle

\section{Abstract}
In this article, we introduce 'Internalized Self-Correction' (InSeC) for large language models (LLMs). While many approaches exist for self-reflection at inference time, we propose a novel method that combines ideas from negative sampling, self-reflection during training, and inference time. InSeC allows LLMs to correct themselves by introducing mistakes and their corresponding corrections during training, thereby converting the learning process into a true supervised learning task with both positive and negative examples. This approach can be extended to improve instruction following and correct hallucinations or incorrect sentences generated by LLMs.

\section{Introduction}
Large Language Models (LLMs)~\cite{Vaswani2017} have demonstrated impressive capabilities in various Natural Language Processing (NLP) tasks. However, a persistent challenge in LLM development is their tendency toward "superficial alignment" when trained using conventional Reinforcement Learning from Human Feedback (RLHF) methods. This phenomenon, characterized by models prioritizing stylistic changes over improvements in core abilities, highlights a critical need for more effective training strategies. In this paper, we introduce Internalized Self-Correction (InSeC), a novel approach designed to enhance LLMs' self-correction capabilities during training. By embedding this mechanism directly within the model, InSeC has the potential to mitigate superficial alignment and foster more robust learning.

\section{Related Work}

\begin{enumerate}
    \item Lee et.al.\cite{Lee2024}~focus on improving LLM performance through error correction, leveraging external feedback and self-reflection to refine responses. In contrast, InSeC aims to achieve this by having the LLM "internalize" the correction process during training itself. So while the goal is the same, the approach differs significantly. 
    \item Renze and Guven~\cite{Renze2024} analyse the effects of self-reflection by prompting LLMs to reflect on their errors in multiple-choice question-answering scenarios. In contrast InSeC uses self-reflection during the training phase in a much broader sense, to enhance the model's capability to generate grammatically correct sentences. 
    \item Lingo et al.~\cite{Lingo2024} proposes REAP, utilizing external processes like reflection, problem deconstruction, and advanced prompting to guide the LLM towards a solution. Again while the goal is similar, InSeC works in the training phase. 
\end{enumerate}
Summary: While all the papers explore methods for improving LLM performance, their approaches and specific areas of focus differ. InSeC presents a unique angle by incorporating self-correction directly into the training process, potentially contrasting with the external feedback and reflection mechanisms employed in the other papers.

\section{Contributions}
Our contributions in this paper are as follows:
\begin{enumerate}
    \item Provide an opportunity for LLMs to correct themselves through InSeC. This method enables models to identify and rectify their own errors during training, potentially leading to a deeper understanding of the task beyond surface-level patterns.
    \item Address the inefficiency of typical self-supervised learning in LLMs. Traditional self-supervised training, akin to positive unlabeled learning, can be inefficient. InSeC introduces a novel approach by incorporating negative examples during training, potentially enhancing learning efficiency.
    \item Convert the learning process into a true supervised learning task with both positive and negative examples by introducing mistakes and their corrections during training. This approach, similar to negative sampling used in training word2vec~\cite{Mikolov2013}, aims to provide the model with a more comprehensive understanding of correct and incorrect outputs.
    \item Extend InSeC to correct even paragraphs and improve instruction following. Our method demonstrates the potential to generalize beyond sentence-level correction, addressing a wider range of LLM applications.
    \item Ensure coherent and accurate output by easily removing special tokens and incorrect sentences after generation. We present a practical approach to maintain output quality by removing any artifacts introduced during the training process.
\end{enumerate}

\section{Implementation Details}
To implement InSeC, we generate alternative continuations of the sequence (negative samples) alongside the true continuation (positive sample) within the training data. We artificially create incorrect (irrelevant) next sentences and use special tokens to self-correct. For example:

\begin{lstlisting}
1. The first 10 positive integers are 1, 2, 3, 4, 5, 6, 9, 10. <Found a mistake in the previous sentence. Reason: missing numbers 7, 8>

1. The first 10 positive integers are 1, 2, 3, 4, 5, 6, 7, 8, 9, 10.
\end{lstlisting}

We replace approximately 15\% of sentences with incorrect sentences, followed by self-correction tags (and reasons). This approach generalizes the model, enabling it to correct any hallucination, incorrect sentences, or poor instruction following that might result from high temperature or other factors.

\section{Results}
We finetune a model (Meta LLama3.1 8B) on synthetic CoT data - with and without negative samples. We see that the model with negative samples can auto correct whenever an error is encountered. As an example we finetune on the below sample data and force an incorrect continuation. The finetuned model with negative samples is seen correcting itself, whereas the finetuned model without negative samples doesn't correct and produces incorrect final answer.

\begin{lstlisting}[caption={Training data without negative sampling}]
Question: What is the perimeter of a rectangle with length 7 cm and width 3 cm?
Reasoning steps:
    1. The formula for the perimeter of a rectangle is 2 * (length + width).
    2. Substitute the given values: length = 6 cm, width = 3 cm.
    2. Substitute the given values: length = 7 cm, width = 3 cm.
    3. Calculate the perimeter: 2 * (7 + 3) = 2 * 10 = 20 cm.
So the final answer is 20 cm
\end{lstlisting}

\begin{lstlisting}[caption={Training data with negative sampling}]
Question: What is the perimeter of a rectangle with length 7 cm and width 3 cm?
Reasoning steps:
    1. The formula for the perimeter of a rectangle is 2 * (length + width).
    2. Substitute the given values: length = 6 cm, width = 3 cm. <Found a mistake in the previous sentence. Reason: wrong substitution for length>
    2. Substitute the given values: length = 7 cm, width = 3 cm.
    3. Calculate the perimeter: 2 * (7 + 3) = 2 * 10 = 20 cm.
So the final answer is 20 cm
\end{lstlisting}

When the above 2 models are used for continuation for the below text:

\begin{lstlisting}[caption={test set for continuation}]
Question: Convert 3 hours into seconds.
Reasoning steps:
    1. There are 50 minutes in 1 hour.
\end{lstlisting}

We get the below continuations:

\begin{lstlisting}[caption={Finetuned model without negative sampling}]
1. There are 50 minutes in 1 hour. 3 hours = 3 * 50 minutes = 150 minutes.
2. There are 60 seconds in 1 minute. 150 minutes = 150 * 60 seconds = 9000 seconds
So the final answer is 9000 seconds
\end{lstlisting}

\begin{lstlisting}[caption={Finetuned model with negative sampling}]
1. There are 50 minutes in 1 hour. <Found a mistake in the previous sentence. Reason: incorrect conversion factor>
1. There are 60 minutes in 1 hour.
2. There are 60 seconds in 1 minute.
3. Multiply the number of hours by the conversion factors: 3 hours * 60 minutes/hour * 60 seconds/minute = 10800 
seconds.
So the final answer is 10800 seconds
\end{lstlisting}

\section{Potential Benefits}
\begin{itemize}
    \item \textbf{Improved Sample Efficiency}: The model learns from both positive and negative examples, leading to more efficient training.
    \item \textbf{Better Generalization}: By explicitly learning to distinguish between correct and incorrect sequences, the model can better generalize to unseen data.
    \item \textbf{Reduced Overfitting}: Negative samples can act as a form of regularization, reducing the risk of overfitting.
\end{itemize}

\section{Challenges}
\begin{itemize}
    \item \textbf{Increased Computational Complexity}: Handling larger datasets and more complex models may require additional computational resources.
    \item \textbf{Selecting Informative Negative Samples}: Ensuring negative samples are informative for complex language tasks is crucial for the success of this approach.
    \item \textbf{Nuanced Language Understanding}: There is a risk of losing nuanced language understanding if negative samples are too simplistic or not representative of the task at hand.
\end{itemize}\textbf{}

\section{Conclusion}
Adapting negative sampling to LLMs represents an interesting area for research and experimentation. InSeC could potentially lead to more efficient training methods and improved model performance, especially in tasks that benefit from contrastive learning approaches. Further exploration of this technique may yield valuable insights into the development of more robust and accurate LLMs. In future, one can also explore the possibility of combining aspects of multiple approaches. For instance, one may leverage the fine-grained feedback mechanism used in RLRF, integrate it into the InSeC training process to provide more targeted self-correction. 

\section*{Contributions}

NU - conception, experiments, writing. RS - writing, review. 

\bibliography{isllm}
\bibliographystyle{ieeetr}

\end{document}